\pdfoutput=1

\documentclass[11pt]{article}

\usepackage{acl}

\usepackage{times}
\usepackage{latexsym}

\usepackage[T1]{fontenc}

\usepackage[utf8]{inputenc}

\usepackage{microtype}
\usepackage{subcaption}
\usepackage{array,multirow,graphicx,booktabs}
\usepackage{makecell}

\newcommand{\STAB}[1]{\begin{tabular}{@{}c@{}}#1\end{tabular}}

\usepackage{amsmath}
\usepackage{mathtools}

\title{Automatic Annotation of Direct Speech in Written French Narratives}

\author{Noé Durandard \\
  Deezer Research \& EPFL \\
  \texttt{\normalsize noe.durandard@epfl.ch} \\\And
   Viet-Anh Tran\\
  Deezer Research\\
  \texttt{\normalsize vatran@deezer.com} \\ \And 
   Gaspard Michel\\
  Deezer Research\\  
\texttt{\normalsize gmichel@deezer.com} \\ \AND
  Elena V. Epure\\
  Deezer Research\\  
\texttt{\normalsize eepure@deezer.com}}

\begin{document}

\maketitle
\begin{abstract}
The automatic annotation of direct speech (AADS) in written text has been often used in computational narrative understanding.
Methods based on either rules or deep neural networks have been explored, in particular for English or German languages.
Yet, for French, our target language, not many works exist.
Our goal is to create a unified framework to design and evaluate AADS models in French.
For this, we consolidated the largest-to-date French narrative dataset annotated with DS per word;
we adapted various baselines for sequence labelling or from AADS in other languages;
and we designed and conducted an extensive evaluation focused on generalisation. 
Results show that the task still requires substantial efforts and emphasise characteristics of each baseline.
Although this framework could be improved, it is a step further to encourage more research on the topic.
\end{abstract}

\section{Introduction}

Prose fiction makes whole worlds emerge.
Authors make use of different strategies to create narratives and convey the \textit{storyworld}.
Novels intertwine narrators' words to build the atmosphere and tell the story, with words stemming from characters inhabiting the fictive world that disclose their personality and depict them directly via dialogues or direct speech (DS)~\citep{james2011art, handbook_narra_2014}.

The development of algorithms to perform the automatic annotation of direct speech (AADS) in written text has been of high interest for literary studies.
This task consists in retrieving lines uttered by the characters of a narrative in contrast to words delivered by the narrator of the story.
One goal of AADS has been to compare fiction works by different authors or stemming from different genres or time periods.
DS was then studied as a literary device carrying specific purposes and disclosing compelling cultural information~\citep{muzny2017dialogism, egbert2020fiction}.
AADS is also central in narrative understanding endeavors.
DS has been then considered as the main realisation of characters, their means to gain volume and depth,
and come alive to the readers.
In this context, AADS is often regarded as a pre-processing step that enables downstream analysis such as DS speaker attribution~\citep{cuesta-lazaro-etal-2022-sea}, that can in turn serve to assemble characters networks~\citep{labatut2019extraction}, or model personas~\citep{sang-etal-2022-tvshowguess}.

\begin{figure}
    \begin{small}
     \centering
     \begin{subfigure}{0.3\textwidth}
         \centering
    \includegraphics[width=\textwidth]{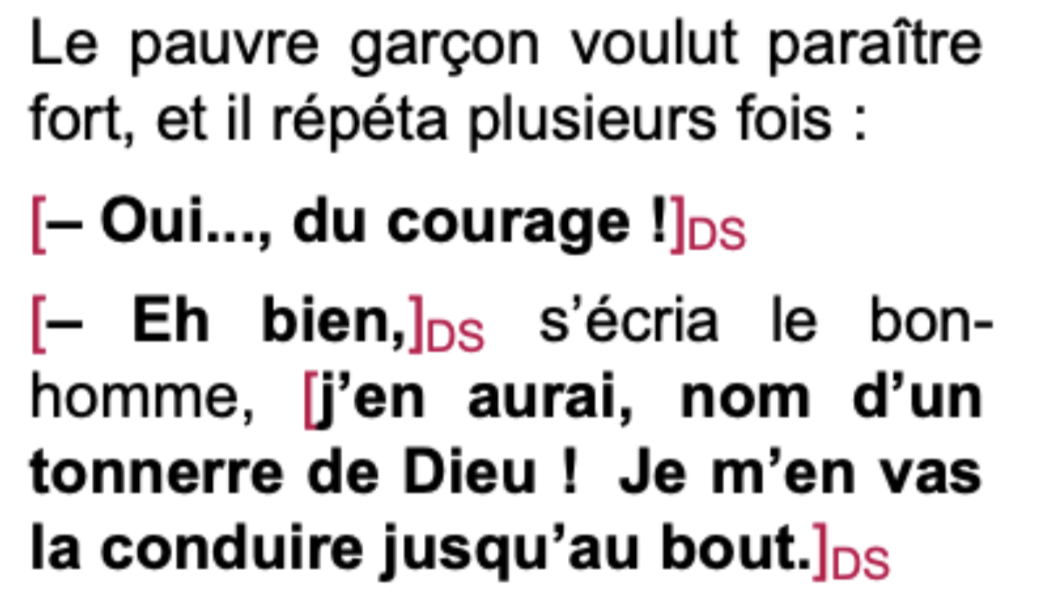}
    \vspace{-6mm}
     \end{subfigure}
\rule{5.5cm}{0.4pt}

     \begin{subfigure}{0.3\textwidth}
         \centering
    \includegraphics[width=\textwidth]{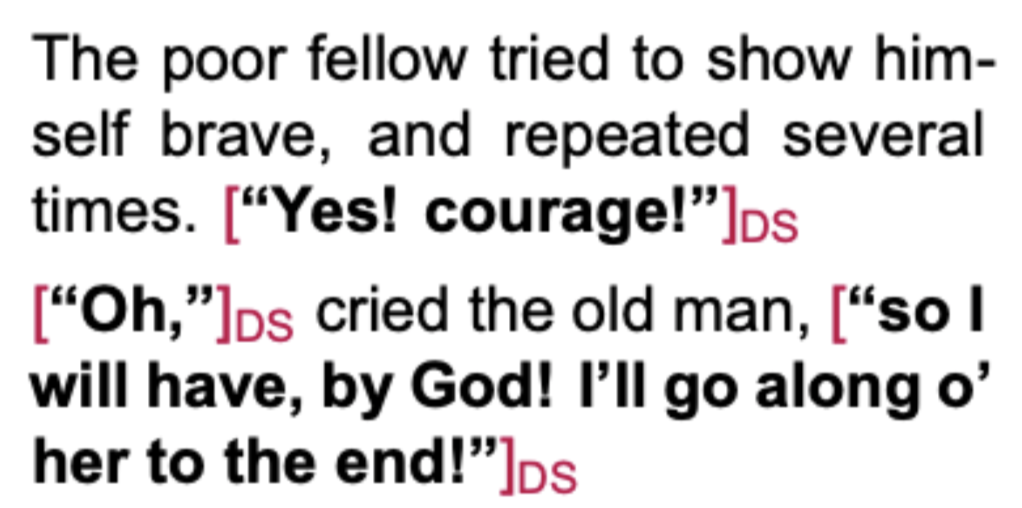}
     \end{subfigure}
        \caption{Excerpts of \textit{Madame Bovary} by Gustave Flaubert (1856). Translation by Eleanor Marx-Aveling.}
        \label{fig:example}
\end{small}  
\end{figure}

AADS has been widely performed for English literature, leveraging strict formatting conventions (e.g. quotes or long dashes) to extract DS through simple regular expression---regex \citep{bamman-etal-2014-bayesian, okeefe-etal-2012-sequence,elson2010automatic}. 
Yet, in other languages, dialogues may be less strictly segregated from narration and typographic conventions can be more flexible. Hence, more complex solutions based on lexical features have been developed, mainly for German \citep{brunner2013automatic, jannidis2018analysing, brunner2020bert}.
These lexical features were either manually defined and exploited with classical machine learning algorithms such as Random Forest \citep{brunner2013automatic}, or were inferred indirectly from text in deep learning frameworks \citep{jannidis2018analysing,brunner2020bert} using Recurrent Neural Networks or language models such as BERT \citep{devlin-etal-2019-bert}.

For other languages, including French, there are very few AADS efforts. 
\citet{schoch2016straight} propose \textit{Straight Talk!}, a corpus of 40 chapters from 19th century French novels annotated per sentence if containing DS or not, and performed binary classification using 81 engineered features. 
The corpus was quite large, but sentences were poorly segmented with a high impact on results;
annotations did not consider \textit{incises} (i.e. narrative breaks within the same DS turn as in Figure \ref{fig:example});
despite a high overall F1-score ($93\%$), some writing styles were very challenging (for instance in homodiegetic narratives, where the narrator is a fully fledged character in the storyworld and may relate the story at the first person). 
In another work, \citet{sini-etal-2018-annotation} adopted a feature engineering approach as well. They combined it with rules to segment and identify paragraphs containing DS, and then to extract \textit{incises} from mixed paragraphs. 
Still, the method was tested on a small corpus, a subset of \textit{SynPaFlex} \cite{sini-etal-2018-synpaflex} with excerpts from only two novels.
Finally, \citet{byszuk-etal-2020-detecting} considered AADS in multilingual settings using BERT, but on an even smaller French corpus.

The goal of the current work is to create an unified framework for designing and evaluating AADS models in French, which in return we hope to encourage more research on the topic\footnote{Code and data are publicly available at \url{https://github.com/deezer/aads_french}}.
Specifically, we address existing limitations on multiple fronts:
\begin{enumerate}
    \item We catalogued and consolidated the largest-to-date dataset of French narratives manually annotated with DS tags at the word level based on $4$ existing corpora. First, we re-annotated \textit{Straight Talk!}~\citep{schoch2016straight} to reach a finer granularity: from sentence to word level. Second, we extended the \textit{SynPaFlex}~\citep{sini-etal-2018-annotation} sparse annotations, initially done on chapter excerpts, to cover the whole chapters. 
    We also incorporated two new corpora as they were: \textit{fr-LitBank}, the French variant of the Multilingual BookNLP project~\citep{githubGitHubLattice8094frlitbank} and an extension of \textit{SynPaFlex}~\citep{sini-etal-2018-annotation} provided by the authors.
    Our dataset is made of $86$ whole chapters ($680K$ annotated tokens), extracted from French novels published during the 19th and 20th centuries.
    
    \item We modelled AADS as a token classification task, which we argue as more suitable for \textit{incises} identification. 
    This approach allowed us to benchmark state-of-the-art sequence labelling models such as French fine-tuned transformers~\citep{martin-etal-2020-camembert} for the first time for AADS.
    We also re-implemented the most popular AADS baselines from other languages to fit French language peculiarities and trained them on our dataset.
    In our selection, we included baselines that did not require extensive manual feature engineering to encourage generalisation over various writing styles.
    
    \item We devised an extensive evaluation covering text with varied formatting quality.
    Apart from traditional token- and span-level strict precision, recall and F1-score metrics~ \citep{yadav-bethard-2018-survey}, we adapted ZoneMap~\citep{galibert_2014}, a metric stemming from page segmentation literature, to our task.
    This allowed us to quantify the effect of various error types made by the models and deepen our understanding of their limitations.
\end{enumerate}

Results show that rule-based baselines using regular expressions remain a good choice when texts are well-formatted.
Deep learning solutions are however more effective and achieve satisfactory results even on narratives with poor formatting quality.
Their most common issue is that they still miss to catch whole DS sequences. 
We also conducted a qualitative analysis to bring insights on the strengths and weaknesses of various models, and defined the directions for future endeavors.

\section{Literature Review} \label{sec:literature}
We further review AADS solutions for any language.

\subsection{Rule-based AADS}
\label{sec:typographicDSR}
While conventions may vary across languages and novels, DS tends to be enclosed within quotation marks (e.g. «...»; “...”), or introduced with long dashes (e.g. —...; --...).
Regarded as pre-processing, simple AADS methods relying on regex with low computational costs are favored~\cite{thomas2012fictional, cunha2004ponctuation}.
The AADS module of BookNLP~\cite{bamman-etal-2014-bayesian}, the reference pipeline developed for computational narrative understanding in English, first determines the most used quotation mark type from a predefined set;
then it tags every passage in between the selected quotation mark pair as DS.
This yields performances around an F1-score of $90\%$ when evaluated as a token-level binary classification task on the LitBank 19th century book corpus~\cite{sims-bamman-2020-measuring}.
Variations of this approach considering more quotation mark types than the most used one are also common \citep{cuesta-lazaro-etal-2022-sea,yoder-etal-2021-fanfictionnlp,byszuk-etal-2020-detecting,okeefe-etal-2012-sequence}.
Almost perfect F1-scores ($96-99\%$) are then reported on various English corpora.

However, when working with heterogeneous corpora, texts with poorer encoding quality (because of Optical Character Recognition errors or changing editing standards over time), or other languages, typographic AADS appears to be limited \citep{byszuk-etal-2020-detecting,muzny2017dialogism}.
For instance, performances on French and German decrease to a F1-score of $92\%$ and down to $65\%$ for Norwegian \cite{byszuk-etal-2020-detecting}.
Similarly, we observe the F1-score decreasing to $77\%$ on a more challenging English corpus \citep{muzny2017dialogism}.

To overcome these issues, more complex rule-based systems that leverage semantic and syntactic cues besides typographic markers have been proposed for English \citep{muzny2017dialogism} and German \citep{tu2019automatic}.
Empirical studies revealing writing style differences between DS and narration \citep{egbert2020fiction} have supported this direction.
The lack of DS markers and the prevalence of \textit{incises} in French literature has also led \citet{sini-etal-2018-annotation} to devise more sophisticated regex based on dependency parsing and Part-of-Speech (POS) tags, yielding an F1-score of $89.1\%$.

\subsection{Machine Learning-based AADS}
\label{sec:lexicaldsr}
With an increasing availability of annotated corpora, AADS based on machine learning has been explored more and more, in particular on German literature~\cite{brunner2013automatic, tu2019automatic, brunner2020bert}.
Works on other languages, such as French~\cite{schoch2016straight} or Swedish~\cite{ek2019distinguishing}, have also emerged, while remaining sparse and isolated. 
ELTeC multilingual initiative \citep{odebrecht_ELTeC_2021} has encouraged the investigation of multilingual approaches too~\cite{byszuk-etal-2020-detecting, kurfali-wiren-2020-zero}.

All these endeavors exploit syntactic and semantic features of DS segments beyond typographic cues, either through feature engineering or by learning features from text with end-to-end deep learning. 
\citet{brunner2013automatic} trained a Random Forest on $80$ syntactic and semantic features extracted at the sentence level from a corpus of 13 short German narratives. 
Her method showed a 3 point improvement compared to the rule-based AADS baseline, though with a large standard deviation ($19\%$). 
This approach was later adapted to French by \citet{schoch2016straight} on a corpus of $40$ book chapters.

In the recent years, the successful application of deep learning to a wide-range of NLP tasks has led to the adoption of these models for AADS too.
\citet{brunner2020bert} proposed to use a BiLSTM-CRF \citep{huang2015bidirectional} on text encoded with Flair \citep{akbik-etal-2018-contextual}, FastText \citep{mikolov2018advances} and a multilingual BERT \citep{devlin-etal-2019-bert}, as well as to fine-tune the German-language BERT \citep{chan-etal-2020-germans} for AADS on German narratives. 
\citet{byszuk-etal-2020-detecting} fine-tuned a multilingual BERT and reported an overall F1-score of 87.3\% at the token level. 
However, the score per language is missing, making it challenging to assess the benefits of the approach for individual cases. 
\citet{kurfali-wiren-2020-zero} adopt a zero-shot framework and remove DS typographic markers from the test corpora. They trained a multilingual BERT on silver-labelled data obtained with regex AADS and report token-level F1-score of 85\% on English, 73\% on Swedish and 64\% on German.

In summary, research dedicated to French remains very sparse and suffers from a lack of comparability because of differences among the studied corpora, task modeling focuses (token vs. sentence classification), or imposed research scenario (without typographic markers, multilingual, zero-shot).

\section{French Narrative Corpora for AADS} 
\label{sec:corpus}

We consolidate a large dataset of French novel excerpts, manually annotated with DS labels at the word level.
Built upon existing endeavors, the final dataset is a compilation of four sub-corpora, individually referred to as \textit{Straight Talk!} (\textit{ST!})~\cite{schoch2016straight}, \textit{SynPaFlex} (\textit{SPF})~\cite{sini-etal-2018-annotation}, an extension of \textit{SynPaFlex} provided to us by the authors (\textit{SB}), and  \textit{fr-LitBank} (\textit{fr-LB})~\cite{githubGitHubLattice8094frlitbank}.
While \textit{fr-LB}, \textit{SPF}, and \textit{SB} have overall good encoding and segmentation quality, \textit{ST!} is poorly formatted with some files lacking line breaks, for instance.

Each sub-corpus contains French novels from public-domain published between 1830 and 1937\footnote{This period is chosen 
because it is copyright-free.}.
It results in an aggregated corpus gathering $86$ chapters extracted from $44$ novels.
The full dataset comprises more than $680K$ words, $8 826$ DS spans which represent $37\%$ of the total tokens.
However, we can observe large variations of DS presence across files (see Appendix \ref{app:corpus}),
from no DS in the excerpt named \textit{madame\_bovary\_première\_9} to 92\% of the words being labelled as DS in \textit{mystères\_de\_paris\_2\_troisième\_16}.
Appendix \ref{app:corpus} shows the excerpts and more dataset details.

The sub-corpora, \textit{fr-LB}, and \textit{SB}, were kept in the form provided by the original works.
In contrast, we modified the ground-truth annotations of \textit{ST!} and \textit{SPF} in order to align them with the other two sub-corpora and the adopted approach to model the problem---binary classification at the token level-- and to exhaustively cover chapters, not only excerpts.
In particular, \textit{ST!} annotations lacked granularity as text segments were labelled as Narration or Mixed (comprising both Narration and DS), so we corrected those. 
As for \textit{SPF}, the annotations were very sparse among the $27$ chapters; hence we extended them to whole chapters.

The re-annotation process was mainly led by one author using Doccano~\cite{doccano}. 
A selection of 5 files were doubly annotated by a co-author to check labeling reliability.
The obtained pairwise Cohen's $\kappa$ score~\cite{cohen_kappa} was 97\%, which is considered almost perfect agreement.
The re-annotated dataset is shared with the code.

\begin{table}
\begin{small}
\centering
\begin{tabular}{@{}lrrrr@{}}
\hline
 & \textbf{\#Fs} & \textbf{\#Toks} & \textbf{\#Sents} & \textbf{DS\%(std)}  \\
\hline

Train & 37 & 333,638 & 22,917 & 40 (30) \\ \hline
Valid & 6 & 67,568 & 5,332 & 34 (19) \\ \hline
Test$_{C}$ & 6 & 59,016 & 3,803 & 29 (19) \\

Test$_{N}$ & 37 & 222,650 & 14,406 & 37 (21) \\ \hline

\hline

\end{tabular}
\caption{Number of files (\#Fs), tokens (\#Toks), and sentences (\#Sents), and \% of DS tokens (with standard deviation) for train, validation, clean test (Test$_{C}$) and noisy test (Test$_{N}$) splits.} \label{tab:corpus-split-spacy}
\end{small}
\end{table}

The dataset is then split into train, validation and test sets.
The files from the three well-formatted sub-corpora (\textit{fr-LB}, \textit{SPF}, \textit{SB}) are randomly divided in order to ensure a proportion of $0.8 / 0.1 / 0.1$ for train, validation, and test, respectively, and that at least one file from each sub-corpus can be found in each split.
Each file can be found in only one split, but we sometimes have files from the same novel present in all splits, especially those originating from the \textit{SPF} sub-corpus (\textit{Les Mystères de Paris} by Eugène Sue and \textit{Madame Bovary} by Gustave Flaubert).
Finally, \textit{ST!} is kept for test only as a challenge dataset.
Indeed, contrary to the other sub-corpora mentioned above, this latter sub-corpus suffers from largely unequal formatting quality across files. 
Some chapters are completely devoid of line break which makes them, wrongly, appear as one unique paragraph, while others exhibit misplaced line breaks, sometimes in the middle of sentences.
\textit{ST!}'s formatting peculiarities make it a good test for generalisation, especially on noisier text. 
This challenging set is also referred to as a noisy test set (Test$_N$) in contrast to the clean test set (Test$_C$) stemming from the split of the three well-formatted sub-corpora —that are also used for training and validation.

Dataset statistics are shown in Table \ref{tab:corpus-split-spacy}.
More details on split composition in terms of files can be found in Appendix \ref{app:corpus}.

\section{Methods}
\label{sec:baselines}
Popular baselines from the two AADS approaches (rule-based and machine learning-based), including those designed for other languages, were modified to fit the characteristics of French. 
AADS was then formalized either as a text matching and extraction task, when using regex, or as a sequence labelling task, when using deep learning models. 
For the latter, the AADS models returned a binary label per token, (\textit{O} / \textit{DS}) as in other related works \citep{brunner2020bert,ek2019distinguishing,jannidis2018analysing}.
While regex has been more common, to our knowledge, this is the most extensive attempt to explore deep learning for AADS in French narratives.

\subsection{Rule-based AADS Baselines} \label{subsec:regex}
We adapted two rule-based systems~\citep{byszuk-etal-2020-detecting,bamman-etal-2014-bayesian} for our framework.

\citet{byszuk-etal-2020-detecting} compiled a list of various quotation marks and dashes used to introduce characters' DS, which we kept the same.
However, we modified the definition of paragraphs, the input to the regex system, to be spans of text until a break line. 
Regular expressions were after applied, as they were, to extract the text enclosed by quotation marks or introduced by a dialogue dash.

In contrast, \citet{bamman-etal-2014-bayesian}'s method was driven by the hypothesis that each text used a single typographic convention for DS.
Thus, they identified the most used quotation mark in the analyzed document from a predefined list.
Then, regex was applied considering only the selected symbols.
To make it applicable to French narratives, we added other types of dialogue cues to the original DS markers list, which we release with the code.

Although \citet{sini-etal-2018-annotation} propose a rule-based algorithm focused on the French language, they relied heavily on crafted syntactic and semantic rules.
Our aim was to avoid extensive manual feature engineering in order to encourage generalisation over various writing styles.
Also, this method was strongly dependent on other external tools for syntactic analysis that introduced further errors too.
Hence, we did not include it in the benchmark. 

\subsection{Deep Learning-based AADS Baselines} \label{subsec:dl_wdsr}
Deep learning-based AADS was modelled as a token classification task, which we considered more suitable for identifying \textit{incises}.
We further discuss how we preprocessed the text in order to maintain a certain degree of contextual coherence for our objective.
Then, we present the two models we included in our benchmark:
1) we adapted the state-of-the-art AADS deep learning model for German~\citep{brunner2020bert} to fit French language peculiarities and re-trained it from scratch
, and 2) we fine-tuned CamemBERT~\citep{martin-etal-2020-camembert} to perform sequence labelling on our dataset.
   
\paragraph{Input Preprocessing.} 
We used spaCy~\cite{honnibal-johnson-2015-improved} to segment text in sentences and each sentence into words and punctuation.

The input length supported by contemporary language or text embedding models is limited.
For instance, BERT~\cite{devlin-etal-2019-bert} accepts a maximum of 512 sub-word tokens, while Flair embeddings~\cite{akbik-etal-2019-flair} initially could handle 512 characters.
This makes them unfitted to represent or produce inferences over whole books, chapters, or even larger paragraphs, which is an important limitation in computational narrative understanding.
However, to preserve a certain degree of coherence within each individual text segment with regard to the DS task,
we implemented an informed split as follows.
Given text in reading direction, a new sentence was added to the existing segment only if the maximum input size $L_C$ was not reached.
Otherwise, the current segment was stored and a new one initialized starting with this last sentence.
We discuss the choice of $L_C$ in Section \ref{sec:experiments}.


\paragraph{Fine-tuned CamemBERT.} To specialize the general linguistic knowledge of the pre-trained language models for a precise purpose---here, to recognize DS, we use fine-tuning.
We work with CamemBERT~\cite{martin-etal-2020-camembert}, one of the reference BERT-like model for French, available in the HuggingFace library~\cite{wolf-etal-2020-transformers}.  
However, as another tokenization of our preprocessed input is performed by CamemBERT, some adjustments were necessary to address out-of-vocabulary limitations and to handle larger sub-word sequences.

First, CamemBERT tokenizer was not be able to project all of the encoded symbols into the model's vocabulary.
This was the case for breaklines as we worked with paragraphs as input, or special space encodings such as "\textbackslash xa0".
We spotted these unknown symbols during a first model tokenization round over the whole set of tokens, initially obtained with spacy, and replaced them with a special token \textit{[UK]}.
Another strategy could have been to remove them but we found these tokens potentially informative for AADS, as text structure cues.

Second, after the CamemBERT tokenization, a sequence of $L_C$ tokens created during preprocessing might result in more sub-word tokens allowed as input.
Similar to BERT, CamemBERT has the input limited to 512 sub-words.
Here, in order to avoid the model automatically truncating the long sequence, the sequence is split in half if it overflows the input limit. 
Thus, it is less likely to have very short sub-sequences and context is evenly shared amongst resulting chunks.
This chunking choice is closely linked to the tested $L_C$ values (up to 512, see section Section \ref{sec:experiments}). 
However, splits are unlikely most of the time, as SpaCy tokens are common French words---most likely represented by one or two sub-words in the model's vocabulary. 

\paragraph{BiLTSM-CRF.}
We adopt the same architecture as in the state-of-the-art AADS model for German proposed by \citet{brunner2020bert}.
Typical for sequence labelling tasks \citep{huang2015bidirectional}, it consists of two bi-directional Long-Short Term Memory (BiLSTM) layers and one Conditional Random Field (CRF) layer.
The model is implemented using the SequenceTagger class of the Flair framework~\cite{akbik-etal-2019-flair}. 
To embed the input, we test multiple options: Flair~\cite{akbik-etal-2019-flair}, FastText~\cite{athiwaratkun-etal-2018-probabilistic}, or Flair and FastText stacked.
Regarding input representation, Flair comes with a native way to handle long sequences, if these are encountered. 
They are chunked and each chunk is pushed to the model while keeping the last hidden state as a new hidden state\footnote{https://github.com/flairNLP/flair/pull/444}.

\section{Experiments}
\label{sec:experiments}

\subsection{Evaluation Metrics} \label{subsec:wdsr-setup-eval}
We assess the performance of models both at the token- and sequence-levels. 
Results are reported overall and per file.
\textit{Token-level} metrics measure the quality of the binary classification per word / token.
Precision, recall and F1 scores are then computed with the scikit-learn library~\cite{scikit-learn}.
\textit{Strict sequence match} (SSM) scores, such as precision, recall and F1 scores, measure the extent to which the predicted DS sequences strictly match the ground-truth ones. 
These are computed with the seqeval library~\cite{seqeval}

We also employ another sequence-level score: \textit{Zone Map Error} (ZME).
This is our custom adaptation of the error computation method originally developed for page segmentation~\cite{galibert_2014}. 
We include ZME because:
1) we wanted to have complementary scores that alleviate SSM's strictness;
2) we aimed to leverage it to get more insights into the quality of the output by studying the impact of various types of errors a model makes.

\begin{table*}
\begin{small}
\centering
\begin{tabular}{r|ccc|ccc}
\hline
& & \textbf{Test$_{C}$} & & & \textbf{Test$_{N}$} & \\
& \textit{Regex} & \textit{BiLSTM-CRF} & \textit{F.CamemBERT} & \textit{Regex} & \textit{BiLSTM-CRF} & \textit{F.CamemBERT} \\
\hline
Tok. F1 & 90 & 83 & \textbf{96} & 47 & 88 & \textbf{93}\\
SSM F1 & 45 & 72 & \textbf{76} & 5.5 & \textbf{30} & 26\\
ZME & 0.23 & 0.41 & \textbf{0.09} & 1.09 & 0.29 & \textbf{0.24} \\
\hline
Av. Tok. F1 & 90 (2.3) & 84 (20) & \textbf{95} (3.8) & 36 (39) & 82 (16) & \textbf{89} (10) \\
Av. SSM F1 & 43 (17) & 71 (22) & \textbf{72} (17) & 5.5 (15) & \textbf{28} (14) & 24 (18) \\
Av. ZME & 0.24 (0.05) & 0.52 (0.85) & \textbf{0.11} (0.08) & 1.13 (1.06) & 0.55 (0.81) & \textbf{0.30} (0.23) \\
\hline
\end{tabular}
\caption{Results, overall (top) and averaged over files (bottom) with standard deviations in parentheses on clean ($C$), and noisy ($N$) test-sets. Best scores are in bold.} \label{tab:whole_results}
\end{small}
\end{table*}

ZME relies on a classification of error types that depends on the overlap between ground-truth and predicted spans.
The overlap can be
\textit{perfect}, \textit{overlapping}, \textit{including} or \textit{included}. 
The error types we could obtain are then: \textit{Match Error} (1-to-1 non-perfect overlapping between ground-truth and predicted spans), 
\textit{Miss} (non-detected ground-truth DS span), 
\textit{False Alarm} (falsely detected DS span), 
\textit{Merge} (several ground-truth DS spans are covered by only one predicted span), 
or \textit{Split} (several predicted spans within a unique ground-truth one). 
The score is also dependent on the span length and the number of correctly classified tokens within a span \citep{galibert_2014}. 
Note that this is an error score, thus it should be minimized.
We present ZME in more detail in Appendix \ref{app:eval-span-zm}.

A final remark is that sequences are not necessarily utterances or turns. 
A single turn can be split into several sequences if it contains \textit{incises} by the narrator.
Reversely, several utterances can be merged in the same sequence if they are not separated by any token labeled as non-DS (O).

\subsection{Experiment Details}

The deep-learning based models were trained using the train split and the best configuration was identified using the validation split.
Only a part of the hyper-parameters were tuned as explained further in this section.
The rule-based baselines do not need training.
However, for space limitation, we report in Section \ref{sec:Results} only the results of the best performing regex baseline on the validation split.
In accordance with the task formalization and most of the existing literature, token-level F1-score was the metric used for model selection, averaged over files to mitigate the influence of longer chapters.

The two rule-based baselines exhibited similar token-level F1-scores on the validation data (over all files): 89\% for BookNLP-inspired method \citep{bamman-etal-2014-bayesian} and 87\% for \citet{byszuk-etal-2020-detecting}'s baseline. 
However, the BookNLP-inspired regex system showed large variance across files and scored $8$ points less than its counterpart baseline adapted from \citep{byszuk-etal-2020-detecting} on the averaged token-level F1-score. Thus, we retained only this latter in further analyses, which we denote \textit{Regex}.

We trained BiLSTM-CRF model for 10 epochs with a batch size of 8 and learning rate set to 0.1. 
After each epoch, performance was assessed on the validation set and the best configuration over epochs was retained.
Regarding the input embeddings, we obtained the largest results for the stacked Flair and FastText, similar to the original work on German AADS~\citep{brunner2020bert}.
We also benchmarked different values (from $64$ to $448$) for the input size $L_C$. 
Both token-level and SSM F1-scores peaked for $L_C=192$ on the validation split, which is the value we keep for test\footnote{The training for each $L_C$ was done with one seed.}. 

We fine-tuned CamemBERT for 3 epochs with a batch size of 8. Similar to the experimental setup of BiLSTM-CRF, we retained the model that yielded the best results on the validation set after any epoch.
We also investigated multiple input size values, $L_C$, from $128$ to $512$.
For each value, training was repeated with 6 different initialisation seeds. 
$L_C=320$ led to the best results.
By manually analysing the sub-word sequences, we noticed that this value corresponded to the maximal input sequence length accepted by the transformer model after the inner preprocessing for length adjustment. 
Indeed, smaller word sequences are likely to result in sub-optimal context use while longer word sequences would more often overflow the input size accepted by the model and be automatically split. 

\section{Results}
\label{sec:Results}

\autoref{tab:whole_results} shows the obtained results, overall (top) and averaged over files (bottom). 
The scores are computed separately on clean (Test$_{C}$) and noisy (Test$_{N}$) data to assess generalization.

\subsection{Performance on well-formatted files} \label{subsec:test_split_performances}
The scores on Test$_{C}$ show that \textit{Regex} is a strong baseline on well-formatted texts, reaching a token-level F1-score of 90\% and a SSM F1-score of 45\% despite its design limitations (e.g. inability to spot \textit{incises}). 
The fine-tuned CamemBERT (\textit{F.CamemBERT}) substantially outperforms \textit{Regex} on all computed metrics, especially on span-level metrics. 
Though \textit{BiLSTM-CRF} has a poorer token-level performance compared to \textit{F.CamemBERT}, it yields a competitive SSM F1-score when averaged over files but with a larger variance. In contrast, \textit{BiLSTM}'s ZME scores are much worse than the \textit{F.CamemBERT}'s ones and are even worse than those of the simple \textit{Regex}. 

\begin{figure}[ht]
     \centering
     \begin{subfigure}[b]{0.4\textwidth}
         \centering
         \includegraphics[width=\textwidth]{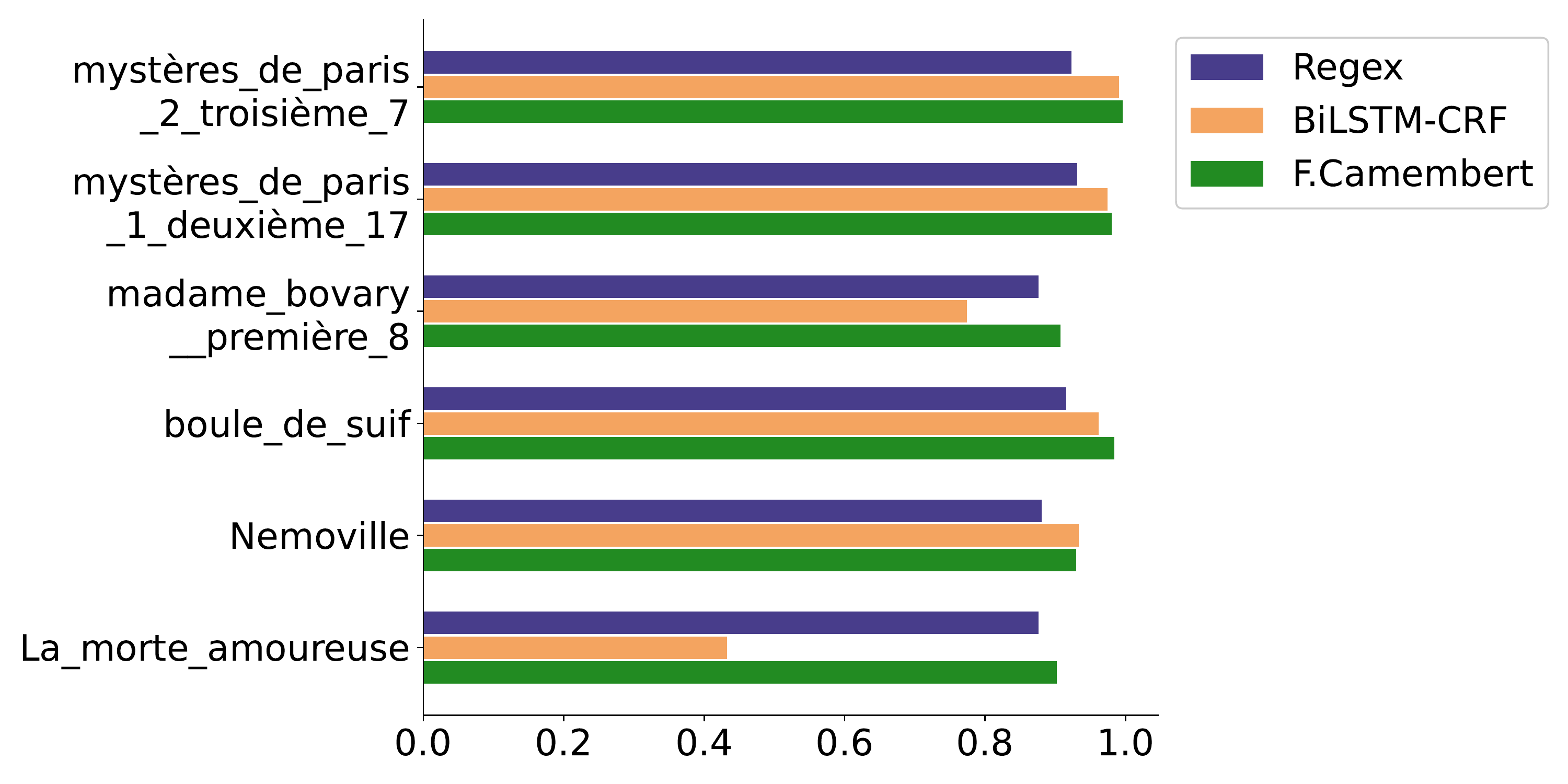}
         \caption{Token-level F1-scores}
         \label{fig:main_per_file_tokf1}
     \end{subfigure}
     \hfill
     \begin{subfigure}[b]{0.4\textwidth}
         \centering
         \includegraphics[width=\textwidth]{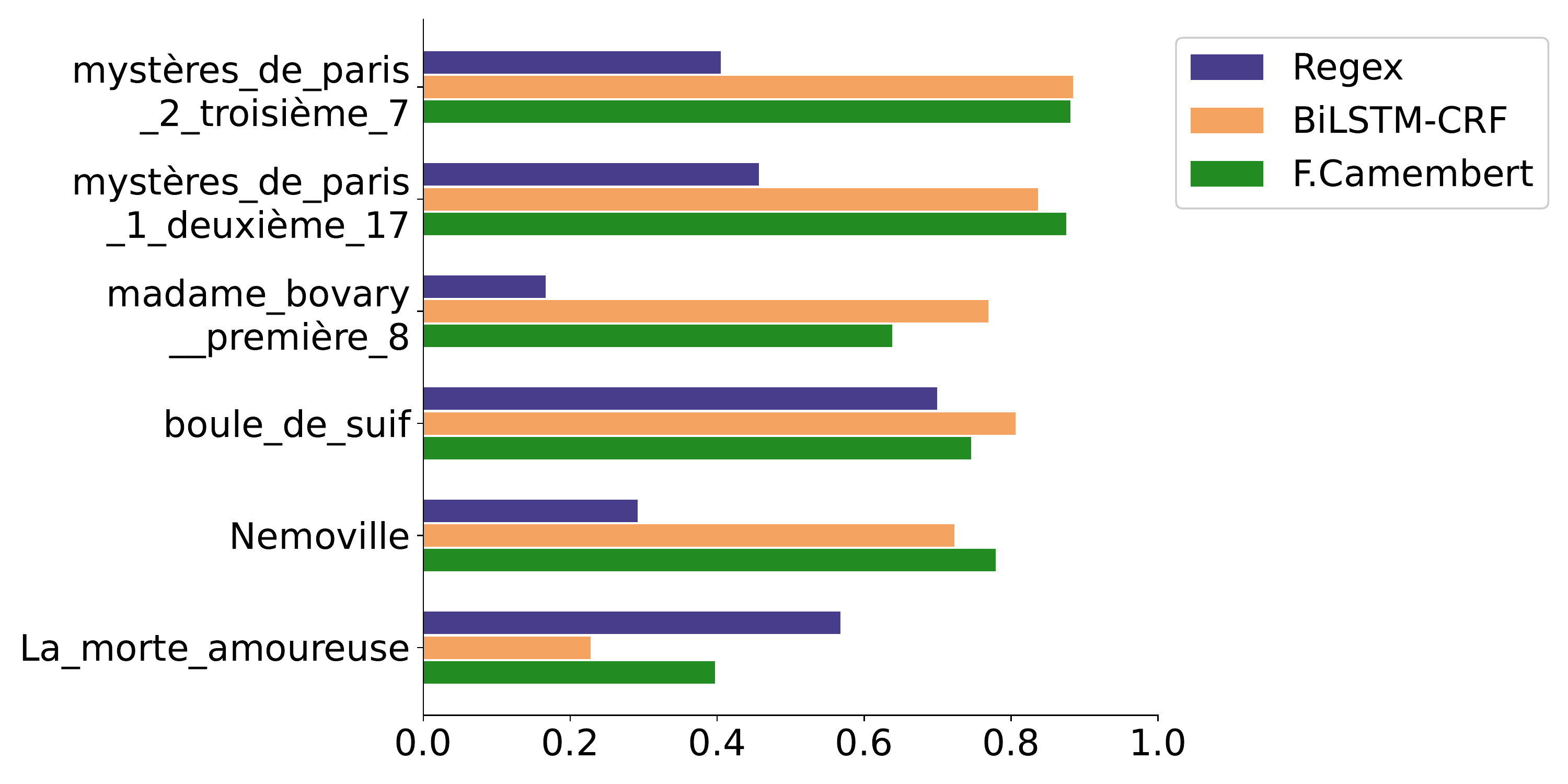}
         \caption{SSM F1-scores}
         \label{fig:main_per_file_ssm1}
     \end{subfigure}
     \hfill
     \begin{subfigure}[b]{0.4\textwidth}
         \centering
         \includegraphics[width=\textwidth]{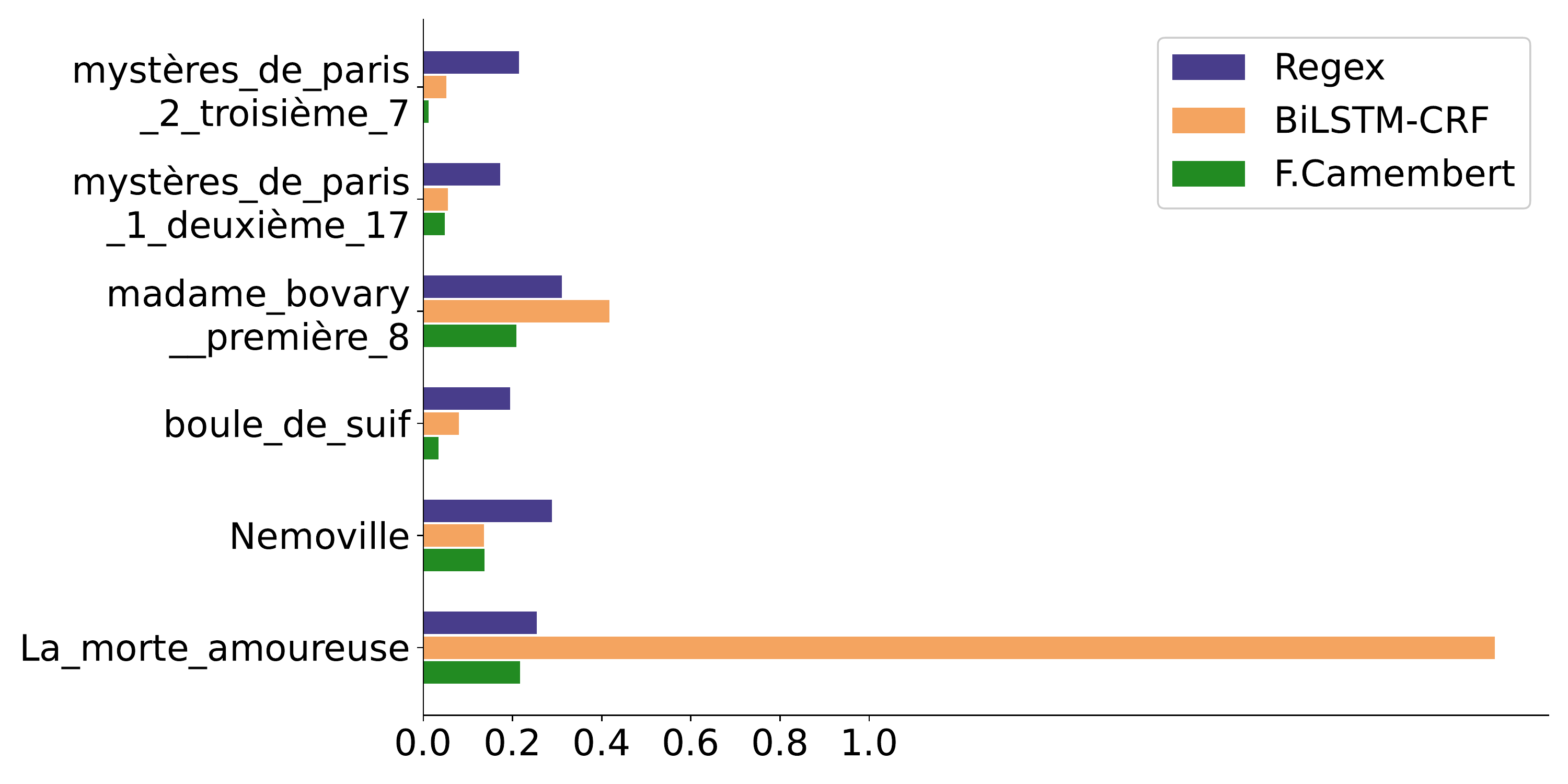}
         \caption{ZME score}
         \label{fig:main_per_file_zme}
     \end{subfigure}
        \caption{Results per file from Test$_{C}$. 
        }
        \label{fig:main_per_file}
\end{figure}

ZME depends on the span length when computing the contribution of each error type (see \autoref{app:eval-span-zm}) and \textit{BiLSTM} appears to make errors concerning longer spans.
Also, as further shown by the performances per file in \autoref{fig:main_per_file}, 
\textit{BiLSTM-CRF} struggles on \textit{La\_morte\_amoureuse}.
This can be, at least partly, explained by the nature of this text. 
The chapter from Théophile Gautier's work is homodiegetic: it is written at the first person ("je") and the character / narrator directly addresses the reader (frequent use of the second person pronoun "vous").
Thus, it could be particularly hard to distinguish DS from narration on this type of text, especially if the model indirectly relies on such cues.
The \textit{F.CamemBERT} seems more robust even in these challenging settings, although it struggles with identifying full spans in this case.

\subsection{Performance on noisy files} \label{subsec:ood_performances}

The results on Test$_{N}$  allows us to get insights on the generalization capabilities of the baselines, in particular when handling low formatting quality.
\textit{Regex} displays poor generalization which was expected given its design and reliance on typographic and formatting cues.
Its token-level F1-score is 53 points less compared to the clean setup in \autoref{tab:whole_results}. 
In fact, \textit{Regex} cannot even detect any DS token on some files as shown in Appendix \ref{app:ood_results}.

In contrast, deep learning-based models are less impacted by changes in formatting quality in terms of token-level F1 scores. 
In this regard, the \textit{F.CamemBERT} remains the best model overall and averaged over files.
\textit{BiLSTM-CRF} shows a better overall token-level F1 score on Test$_{N}$ than on Test$_{C}$ (88\% vs. 83\%).
As shown in Appendices \ref{app:corpus} and \ref{app:ood_results}, it is linked to the model obtaining very good scores on chapters with many DS tokens.

Moreover, the deep learning models are much better than \textit{Regex} on the span-level metrics.
\textit{BiLSTM-CRF} is slightly more competitive than \textit{F.CamemBERT}, but the average over files SSM F1-scores are not significantly different. 
Indeed, as emphasized by the results per file in Appendix \ref{app:ood_results}, the performance is chapter-dependent.
While \textit{F.CamemBERT} consistently outperforms the other baselines on token-level F1-score on all files, \textit{BiLSTM-CRF} is better at recognizing DS spans in about 22 out of 37 files (i.e. $60\%$ of the time). 
However, we could notice again that the \textit{BiLSTM-CRF}'s ZME scores are quite large but more stable than \textit{F.CamemBERT} when the test set moves from clean ($C$) to noisy ($N$) (0.02 vs. 0.19 between the two setups).
In spite of that, \textit{F.CamemBERT} clearly appears as the best-performing model in both cases. 


\begin{table*}
\begin{small}
\centering
\begin{tabular}{r|cccc}
\hline
Error Type & \textbf{Miss} & \textbf{False Alarm} & \textbf{Split} & \textbf{Merge} \\
\hline
\textit{Regex} & 1139.8 (1153.8) & 63.2 (117.7) & 208.0 (545.8) & 633.4 (2318.0) \\
\textit{BiLSTM-CRF} & 55.6 (93.2) & 144.9 (212.4) & 230.1 (369.5) & 21.8 (35.1) \\
\textit{F.CamemBERT} & 1.7 (4.1) & 106.7 (140.0) & 342.0 (566.6) & 12.3 (63.6) \\
\hline
\end{tabular}
\caption{Contribution of various error types to the ZME score, averaged across file, on Test$_{N}$. Standard deviation is reported in parentheses.}
\label{tab:zme_groups_stats}
\end{small}
\end{table*}

\subsection{Qualitative analysis and Discussion}\label{subsec:qualitative}
We conducted a qualitative analysis by checking the detailed contribution of each ZME error type for all models and by manually comparing a selection of files\footnote{Files from Test$_{C}$: \textit{La\_morte\_amoureuse} and \textit{mystères\_de\_paris\_2\_troisième\_7}. Files from Test$_{N}$: \textit{rd0002\_1}, \textit{rd0724\_0}, and \textit{rd0367\_1}.} with their corresponding predictions. 
Table \ref{tab:zme_groups_stats} reveals interesting differences (despite a lack of statistical significance) between \textit{BiLSTM-CRF} and \textit{F.CamemBERT} on Test$_{N}$. 
While \textit{BiLSTM-CRF} exhibits more \textit{Miss}, \textit{False Alarm} and \textit{Merge} error type contributions to ZME, \textit{F.CamemBERT}'s ZME score is more impacted by \textit{Split} errors.
The manual investigation of the selected files showed that both deep learning-based models identified much better \textit{incises} than \textit{Regex}.
This is also consistent with the much lower \textit{Merge} ($21.8$ and $12.3$ vs. $633.4$). 
Nonetheless, other semantic, syntactic or lexical cues seemed to mislead these models. 


On the one hand, \textit{BiLSTM-CRF} seemed to systematically classify parts of text written at the first person ("je") as DS, which makes it especially unfitted for homodiegetic novels (hence the low performance on \textit{La\_ morte\_ amoureuse}). 
The punctuation seemed to be a strong cue for the model as it tended to classify sentences with exclamation or interrogation marks as DS.
Then, \textit{BiLSTM-CRF} could not handle long and uninterrupted DS paragraphs. 
These long DS spans often share registries or production strategies similar to narration~\citep{egbert2020fiction}, such as the use of past tense or descriptions, which likely misled the model. 

\begin{figure}
\begin{small}
    \centering
    \includegraphics[width=0.3\textwidth]{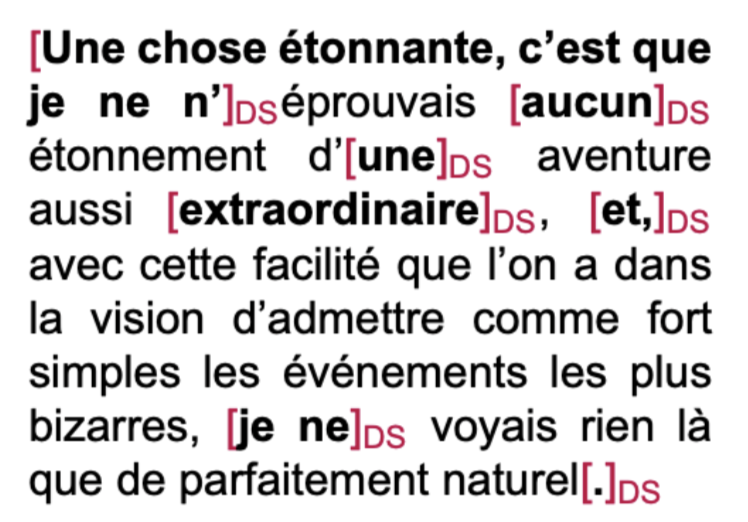}
    \vspace{-1mm}
    \caption{Narration excerpt of La Morte Amoureuse by Théophile Gautier (1836) annotated by \textit{F.CamemBERT}.}
    \label{fig:transformer_short_seq_error}
\end{small}
\end{figure}

On the other hand, the manual analysis showed that \textit{F.CamemBERT} appeared better at identifying long DS spans or at classifying homodiegetic narration.
However, this model bears other weaknesses.
For instance, proper noun phrases seemed to be systematically classified as non-DS.
Another common error was the miss-classification as non-DS of \textit{[UK]} tokens in files using unrecognized non-breaking spaces (e.g. "$\backslash$xa0") after quotation marks.
Plus, the model regularly produced chains of alternating labels on very short groups of words as in~\autoref{fig:transformer_short_seq_error}.
These aspects correlated with the high contribution to ZME from \textit{False Alarm} and \textit{Split} error types.

\textcolor{black}{
Finally, these observations also motivated a final post-processing AADS experiment.
A simple heuristic is used a posteriori to hinder incoherent predictions that mix different narrative levels within the same segment of a sentence.
The correction of the predicted labels post-model using a majority vote per clause lead to significant improvements on sequence-level metrics for both of the deep learning-based models.
Indeed, in all settings --overall and averaged on both clean and noisy files-- \textit{F.CamemBERT}'s SSM F1 scores gained from 5 to 8 points. The performances of the \textit{BiLSTM-CRF} model are only slightly impacted on Test$_C$ but its SSM F1 scores gained in average 5 points on Test$_N$. After this post-processing step, \textit{F.CamemBERT} shows weaker performances than \textit{BiLSTM-CRF} only on SSM F1 scores averaged over files on Test$_N$.
Details of this clause-consistent post-processing step, as well as ensuing results, are reported in \autoref{sec:clause-consistent}.
Altogether the different experiments tend to show that \textit{F.CamemBERT} is the most promising model for AADS, when computational resources and ground-truth annotation are available for training.
}


\section{Limitations}
\label{limitations}
The current framework bears several limitations. 

First, although a common strategy in the related literature \citep{brunner2020bert,ek2019distinguishing,jannidis2018analysing} which we also adopted, the binary annotation at the token-level is limiting. 
With this schema, the focus is not on speakers' utterances or turns, but on DS sequences.
A subsequent issue is that consecutive turns by different characters are considered as one DS sequence if there is no "O" labeled tokens between them.
One solution could have been to mark the start and end of a DS turn while paying attention to handle imbricated narration (ie. \textit{incises}).
However, this would have required significant more re-annotation efforts, which we left for a future research cycle within the proposed framework.
 
Second, because of copyright issues the corpus contains excerpts exclusively from a specific period, 1830-1937.
Thus, the models were trained and tested on a specific type of literature and may not generalize well to other forms of narratives, in particular modern and contemporary.
In this direction, the curation of the test corpus could benefit from more literary insights considering that the evaluation showed high variance of the performance over chapters. 
This could help to better determine the application scope of the models, and which kind of narratives require further work.

With regard to the deep neural network baselines, we did not perform an extensive parameter search and model optimisation.
This could have further improved the results.
However, performances on recognizing full DS spans were clearly lower than token-level metrics, which had most likely other causes.
Regarding the evaluation, although we adopted ZME scores from page segmentation to have more qualitative insights, there are still other aspects we have not quantified and could be particularly relevant.
For instance, does the model tend to miss the beginning, the end or some other specific parts of a DS sequence? 
We tried to capture some of these phenomena through our manual analysis, but it is challenging to apply it at scale without introducing methods to  automatically compute metrics.

\section{Conclusion}
\label{sec:Conclusion}
We have presented an unified framework to design and evaluate AADS in written French narratives.
To our knowledge, this is the largest AADS study to date in French.
We consolidated a large dataset annotated per word. 
Then, we benchmarked two families of baselines, rule and deep learning-based, using as inspiration AADS advances in other languages (German and English).
We designed an evaluation which focuses on generalization and on learning about the advantages and weaknesses of each baseline.
Results show that rule-based systems work well on bounded DS conventions (quotation marks) in clean text.
Other DS formats, \textit{incises}, and poorly formatted files pose many problems.
Deep learning baselines prove to be far more robust, reaching token-level F1-scores up to 95\%, but with large variance across files.
Yet, recognizing full spans of DS is still challenging, even when texts have good formatting quality.

While for macro analyses in literary studies, imperfect AADS may be sufficient, some use-cases require almost perfect performance when recognizing DS spans (e.g. audiobook generation from text).
If a more thorough parameter optimization might help, our qualitative analysis conveys that performance gain should be instead sought by integrating domain knowledge into the models---without feature over-engineering though. 
Studying the models' performances after the removal of typographic cues could lead to other insights on how to increase robustness.
Multilingual language models and existing AADS corpora could be also exploited for French.
Another needed step would be to transition 
to identifying full DS turns and their corresponding speakers, with the implied manual re-annotation efforts.

\bibliography{anthology,custom}

\clearpage

\appendix
\section{Corpus Details}
\label{app:corpus}
Corpus details (file names, authors, publication years and DS percentages per excerpt) are given in Table \ref{tab:corpus-main-details} for the clean ($C$) corpus and in Table \ref{tab:corpus-ood-details} for the noisy ($N$) corpus.
\begin{table}\centering\begin{scriptsize}
\begin{tabular}{@{}lccc@{}}
\toprule
\textbf{File Name} & \textbf{Author} & \textbf{Year} & \textbf{\%DS} \\
\midrule

\scriptsize{rd0571\_0} & Balzac & 1841 & 29 \\
\scriptsize{rd0571\_1} & Balzac & 1841  & 31 \\
\scriptsize{rd0127\_1} & Sue & 1842 & 83 \\
\scriptsize{rd0444\_0} & Sand & 1845 & 80 \\
\scriptsize{rd0444\_1} & Sand & 1845 & 47 \\
\scriptsize{rd0724\_0} & Dumas & 1849 & 2 \\
\scriptsize{rd0724\_1} & Dumas & 1849 & 16 \\
\scriptsize{rd0002\_0} & Aurevilly & 1852 & 18 \\
\scriptsize{rd0002\_1} & Aurevilly & 1852 & 32 \\
\scriptsize{rd0623\_0} & FevalPP & 1852 & 20 \\
\scriptsize{rd0623\_1} &  FevalPP & 1852 & 23 \\
\scriptsize{rd0616\_0} & FevalPP & 1856 & 29 \\
\scriptsize{rd0616\_1} &  FevalPP & 1856  & 49 \\
\scriptsize{rd1169\_0} & Ponson & 1859 & 64 \\
\scriptsize{rd1169\_1} &  Ponson & 1859  & 33 \\
\scriptsize{rd1160\_0} & Ponson & 1860 & 10 \\
\scriptsize{rd1160\_1} &  Ponson & 1860 & 55 \\
\scriptsize{rd0730\_0} & About & 1862 & 34 \\
\scriptsize{rd0730\_1} & About & 1862 & 11 \\
\scriptsize{rd0305\_0} & Aimard & 1868 & 81 \\
\scriptsize{rd0305\_1} &  Aimard & 1868  & 44 \\
\scriptsize{rd1029} & Gaboriau & 1867 & 46 \\
\scriptsize{rd1152\_0} & Gaboria  & 1873 & 52 \\
\scriptsize{rd1152\_1} & Gaboria  & 1873  & 45 \\
\scriptsize{rd0061\_1} & Zola & 1873 & 15 \\
\scriptsize{rd0061\_2} & Zola & 1873 & 23 \\
\scriptsize{rd0014\_0} & Verne & 1877 & 3 \\
\scriptsize{rd0014\_1} &  Verne & 1877  & 53 \\
\scriptsize{rd0367\_0} & Gouraud & 1882 & 23 \\ 
\scriptsize{rd0367\_1} &  Gouraud & 1882 & 30 \\
\scriptsize{rd0407\_0} & Malot & 1878 & 55 \\
\scriptsize{rd0407\_1} &  Malot & 1878 & 36 \\
\scriptsize{rd0656\_0} & Ohnet & 1885 & 46 \\
\scriptsize{rd0656\_1} &  Ohnet & 1885  & 31 \\
\scriptsize{rd0423\_0} & Mary & 1886 & 55 \\
\scriptsize{rd0423\_1} &  Mary & 1886  & 22 \\
\scriptsize{rd1009} & Boisgobey & 1888 & 67 \\

\bottomrule
\end{tabular}\end{scriptsize}
\caption{\textit{ST!} corpus details used as noisy test-set.} \label{tab:corpus-ood-details}
\end{table}

\begin{table*}\centering\begin{scriptsize}
\begin{tabular}{@{}cclllr@{}}
\toprule
&& \textbf{File Name} & \textbf{Author} & \textbf{Year} & \textbf{\%DS} \\
\midrule

\multirow{60}{*}{\STAB{\rotatebox[origin=c]{90}{\textbf{TRAIN}}}}
& \multirow{18}{*}{\STAB{\rotatebox[origin=c]{90}{\textit{fr-LB}}}}

    & \scriptsize{Sarrasine} & Honoré de Balzac & 1830 & 22 \\ \\
    && \scriptsize{Pauline} & George Sand  & 1841 & 28 \\ \\
    && \scriptsize{Madame\_de\_Hautefort} & V. Cousin  & 1856 & 12 \\ \\
    && \scriptsize{Le\_capitaine\_Fracasse} & Théophile Gautier & 1863 & 5 \\ \\
    && \scriptsize{Le\_ventre\_de\_Paris} & Émile Zola & 1873 & 15 \\ \\
    && \scriptsize{Bouvard\_et\_Pecuchet} & Gustave Flaubert & 1881 & 4 \\ \\
    && \scriptsize{Mademoiselle\_Fifi\_nouveaux\_contes-1} & Guy de Maupassant & 1883 & 17 \\ \\
    && \scriptsize{Mademoiselle\_Fifi\_nouveaux\_contes-3} & Guy de Maupassant & 1883 & 19 \\ \\
    && \scriptsize{Rosalie\_de\_Constant\_sa\_famille\_et\_ses\_amis} & Lucie Achard & 1901 & 68 \\ \\
    && \scriptsize{elisabeth\_Seton} & Laure Conan & 1903 & 24 \\ \\
    && \scriptsize{Jean-Christophe-1} & Romain Rolland & 1912 & 10 \\ \\
    && \scriptsize{Jean-Christophe-2} & Romain Rolland & 1912 & 3 \\ \\
    && \scriptsize{Douce\_Lumiere} & Marguerite Audoux & 1927 & 11 \\ \\
    && \scriptsize{De\_la\_ville\_au\_moulin} & Marguerite Audoux & 1937 & 9 \\ \\
    
\cmidrule{2-6}
& \multirow{18}{*}{\STAB{\rotatebox[origin=c]{90}{\textit{SPF}}}}

    & \scriptsize{mystères\_de\_paris\_1\_première\_1} & Eugène Sue & 1843 & 36 \\
    && \scriptsize{mystères\_de\_paris\_1\_première\_3} &  &  & 86 \\
    && \scriptsize{mystères\_de\_paris\_1\_première\_4} &  &  & 73 \\
    && \scriptsize{mystères\_de\_paris\_1\_première\_11} &  &  & 79 \\
    && \scriptsize{mystères\_de\_paris\_1\_première\_15} &  &  & 64 \\
    && \scriptsize{mystères\_de\_paris\_1\_deuxième\_1} &  &  & 65 \\
    && \scriptsize{mystères\_de\_paris\_1\_deuxième\_7} &  &  & 95 \\
    && \scriptsize{mystères\_de\_paris\_2\_troisième\_3} &  &  & 71 \\
    && \scriptsize{mystères\_de\_paris\_2\_troisième\_6} &  &  & 83 \\
    && \scriptsize{mystères\_de\_paris\_2\_troisième\_20} &  &  & 50 \\
    && \scriptsize{mystères\_de\_paris\_2\_troisième\_16} &  &  & 92 \\
    && \scriptsize{mystères\_de\_paris\_2\_quatrième\_4} &  &  & 86 \\
    && \scriptsize{mystères\_de\_paris\_2\_quatrième\_12} &  &  & 77 \\
    && \scriptsize{mystères\_de\_paris\_2\_quatrième\_15} &  &  & 59 \\
    && \scriptsize{mystères\_de\_paris\_2\_quatrième\_19} &  &  & 46 \\ \\
    && \scriptsize{madame\_bovary\_\_première\_1} & Gustave Flaubert & 1857 & 3 \\
    && \scriptsize{madame\_bovary\_\_première\_3} &  &  & 19 \\
    && \scriptsize{madame\_bovary\_\_première\_9} &  &  & 0 \\
    && \scriptsize{madame\_bovary\_\_deuxième\_1} &  &  & 27 \\
    && \scriptsize{madame\_bovary\_\_deuxième\_9} &  &  & 24 \\
    && \scriptsize{madame\_bovary\_\_troisième\_10} &  &  & 13 \\
    
\cmidrule{2-6}
& \multirow{3}{*}{\STAB{\rotatebox[origin=c]{90}{\textit{SB}}}}

    & \scriptsize{contesse\_menager} & Leblanc & 1924 & 49 \\ \\
    && \scriptsize{cousinebette\_cecile} & Balzac & 1847 & 41 \\ \\

\midrule
\multirow{12}{*}{\STAB{\rotatebox[origin=c]{90}{\textbf{VALIDATION}}}}

& \multirow{3}{*}{\STAB{\rotatebox[origin=c]{90}{\textit{fr-LB}}}}

    & \scriptsize{export} & Guy de Maupassant & 1898 & 11 \\ \\
    && \scriptsize{Le\_diable\_au\_corps} & Raymond Radiguet & 1923 & 6 \\ \\

\cmidrule{2-6}    
& \multirow{3}{*}{\STAB{\rotatebox[origin=c]{90}{\textit{SPF}}}}
    & \scriptsize{mystères\_de\_paris\_1\_première\_17} & Eugène Sue & 1843 & 48 \\ \\
    && \scriptsize{mystères\_de\_paris\_1\_deuxième\_20} & Gustave Flaubert & 1857  & 52 \\
    && \scriptsize{mystères\_de\_paris\_2\_troisième\_11} &  &  & 54 \\
    
\cmidrule{2-6}    
& \multirow{2}{*}{\STAB{\rotatebox[origin=c]{90}{\textit{SB}}}}
    & \scriptsize{notaire\_dedier} & About & 1862 & 39 \\ \\

\midrule
\multirow{12}{*}{\STAB{\rotatebox[origin=c]{90}{\textbf{TEST}}}}

& \multirow{3}{*}{\STAB{\rotatebox[origin=c]{90}{\textit{fr-LB}}}}
    & \scriptsize{La\_morte\_amoureuse} & Théophile Gautier & 1836 & 14 \\ \\
    && \scriptsize{Nemoville} & Adèle Bourgeois & 1917 & 32 \\ \\
    
\cmidrule{2-6}    
& \multirow{4}{*}{\STAB{\rotatebox[origin=c]{90}{\textit{SPF}}}}
    & \scriptsize{mystères\_de\_paris\_1\_deuxième\_17} & Eugène Sue & 1843 & 51 \\ 
    && \scriptsize{mystères\_de\_paris\_2\_troisième\_7}  & & & 60 \\ \\
    && \scriptsize{madame\_bovary\_\_première\_8} & Gustave Flaubert & 1857 & 6 \\ \\
    
\cmidrule{2-6}    
& \multirow{2}{*}{\STAB{\rotatebox[origin=c]{90}{\textit{SB}}}}
    & \scriptsize{boule\_de\_suif} & Maupassant & 1870 & 15 \\ \\

\bottomrule
\end{tabular}\end{scriptsize} 
\caption{Well-formatted clean corpus details shown per file and splits (train, validation, test).}\label{tab:corpus-main-details}
\end{table*}

\section{ZoneMap} \label{app:eval-span-zm}

The ZoneMap Error metric~\cite{galibert_2014} was originally developed for page segmentation.
ZoneMap offers a configurable way to compute area segmentation errors based on a typology of possible errors. 

Let $N_T$, $N_P$ be respectively the number of positive (here, DS) spans from the ground truth, and from the model's predictions. The corresponding sets can respectively be written as $\{s_i\}_{i=1}^{N_T}$ and $\{\Tilde{s_j}\}_{j=1}^{N_P}$. The length of a span $s_k$ (given in terms of tokens) is written as $|s_k|$. Ground truth and predicted spans are grouped according to rules described further into $N$ groups $G_k$, $k=1,...,N$.
Then, the error score attributed to the model is given by:
\begin{align}
      E_\mathrm{ZM} &= \frac{\sum_{k=1}^{N}E(G_k)}{\sum_{i=1}^{N_T}|s_i|
      } \label{eq:zme_tot}\\
    \text{where }E(G_k)&=(1-\alpha_C)E_S(G_k)+\alpha_C E_C(G_k) 
\label{eq:zme_group}
\end{align}
with $\alpha_C\in[0,1]$.
$ E(G_k)$ is a linear interpolation of the segmentation error rate $E_S$ and the classification error rate $E_C$ within group $k$. 
Both error types can be defined purposely to penalize the model differently depending on the group type of $G_k$. Groups' constructions, types and compositions are defined below.

Groups are constructed based on a \textit{link force} between true and predicted spans computed as: 
\begin{equation} \label{eq:zme_linkf}
\begin{aligned}
    f_{i,j} \coloneqq f(s_i, \tilde{s_j}) = \left(\frac{|s_i\cap \tilde{s_j}|}{|s_i|}\right)^2+\left(\frac{|s_i\cap \tilde{s_j}|}{|\tilde{s_j}|}\right)^2\\
    i\in\{1,\cdots, N_T\},\ j\in\{1,\cdots, N_P\}
\end{aligned}
\end{equation}
Non-zero links are then sorted in descending order and areas are combined into groups incrementally according to one rule: if adding a new area to a group leads to the situation where a group contains multiple ground truth or predicted areas, then such an area is not added to the group in question.
This process ultimately results in five types of groups:
\begin{enumerate}
    \item \textit{Match}: one ground truth area overlaps with one predicted area and none of them overlap with other predicted or ground truth areas (even if the covered areas are not aligned).
    \item \textit{\textit{Miss}}: one ground truth area is not covered at all by any predicted area.
    \item \textit{False Alarm}: one predicted area is not covered at all by any ground truth area.
    \item \textit{Split}: one ground truth area is covered by at least two predicted areas.
    \item \textit{Merge}: one predicted area is covered by at least two ground truth areas.
\end{enumerate}

Considering the nature of the AADS task as a binary classification, spans will be used instead of areas and classification error rates will be omitted further (set $\alpha_C=0$). 

For both \textit{Miss} and \textit{False Alarm} groups, the segmentation error rate is strictly proportional to their length: 
if $G_k=\{s_i\}$, respectively $G_k=\{\tilde{s_j}\}$; the group contribution to the Zone Map error is $E(G_k)=|s_i|$, respectively $E(G_k)=|\tilde{s_j}|$. 
For \textit{Match} groups, the group error is proportional to the number of non-overlapping tokens: if $G_k=\{s_i, \tilde{s_j}\}$, then $E(G_k)=|s_i\cup\tilde{s_j}|-|s_i\cap\tilde{s_j}|$, so that $E(G_k)=0$ for strict span matches.

Finally, \textit{Split} and \textit{Merge} groups are divided into sub-zones that are in turn classified as strict \textit{Match} and \textit{Miss} or \textit{False Alarm}. 
\textit{Miss} and \textit{False Alarm} sub-zones contribute to the error like \textit{Miss} or \textit{False Alarm} groups (strictly proportionally to the length of the sub-zones). 
In contrast, the largest \textit{Match} sub-zone is not counted as an error and does not contribute to $E_\mathrm{ZM}$, while the remaining \textit{Match} sub-zones are partially counted as errors. 
Their contribution to the segmentation error rate is proportional to their length, an introduced \textit{\textit{Merge}}/\textit{\textit{Split}} mitigating parameter $\alpha_{MS}\in[0,1]$ and the relative number of split, respectively merged sub-zones. 
Given a \textit{Split} group $G_k=\{s_i, \{s_{j,l}\}_{l=1}^{n_k}\}$, the group is sub-divided into $n_k$ strict match sub-zones $\{z_m\}_{m=1}^{n_k}$ and $\overline{n_k}\in\{n_k-1,n_k,n_k+1\}$ non-overlapping spans $\{\overline{z_m}\}_{m=1}^{\overline{n_k}}$. 
The segmentation error rate of such group would then be the sum of non-detected tokens $E_{F}(G_k)=\sum_{m=1}^{\overline{n_k}}|\overline{z_m}|$ and split penalization $E_{T}=\alpha_{MS}\left(\sum_{m=2}^{n_k}|z_m|\right)\frac{n_k-1}{n_k}$. Those formula can then be rewritten in terms of original spans as:

\begin{align}
E_F(G_k) &= \left| s_i\cup\left(\cup_{l=1}^{n_k}\tilde{s_{j,l}}\right) \right| - \left|s_i\cap \left(\cup_{l=1}^{n_k}\tilde{s_{j,l}}\right) \right| \\
E_T(G_k) &= \alpha_{MS} V_{i,j}  \frac{n_k-1}{n_k} \\
V_{i,j}&=|s_i\cap \left(\cup_{l=1}^{n_k}\tilde{s_{j,l}}\right)|-\max_{l\in\{1,\cdots,n_k\}} |s_i\cap\tilde{s_{j,l}}|
\end{align}
and $E(G_k) = E_F(G_k)+E_T(G_k)$. \textit{Merge} groups error contribution is computed comparably.

\begin{figure}
    \centering
\includegraphics[width=.45\textwidth]{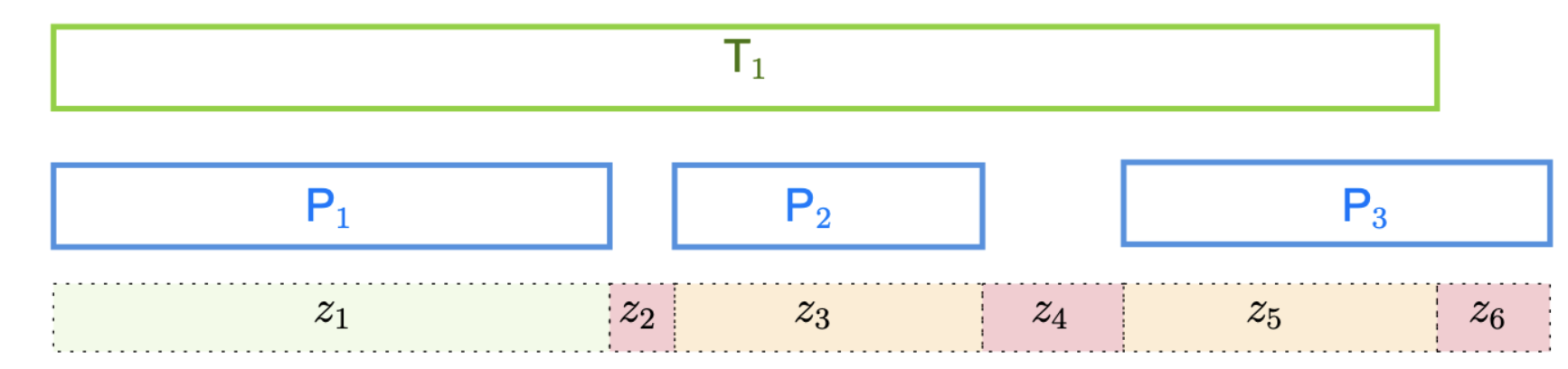}
    \caption{Split configuration type. Ground Truth span $\mathrm{T}_1$ is grouped together with Predicted spans $\mathrm{P}_1$, $\mathrm{P}_2$ and $\mathrm{P}_3$. Sub-zones $z_2$, $z_4$ and $z_6$ contribute fully to the error as \textit{Miss} and \textit{False Alarm}, while $z_3$ and $z_5$ have a mitigated contribution (factored by $(2\alpha_{MS})/3$), and $z_1$ doesn't contribute to the error.}
    \label{fig:zme_split}
\end{figure}

\section{Out-of-distribution results per file}
\label{app:ood_results}
\begin{figure*}
\begin{small}
     \centering
     \begin{subfigure}[b]{0.4\textwidth}
         \centering
         \includegraphics[width=\textwidth]{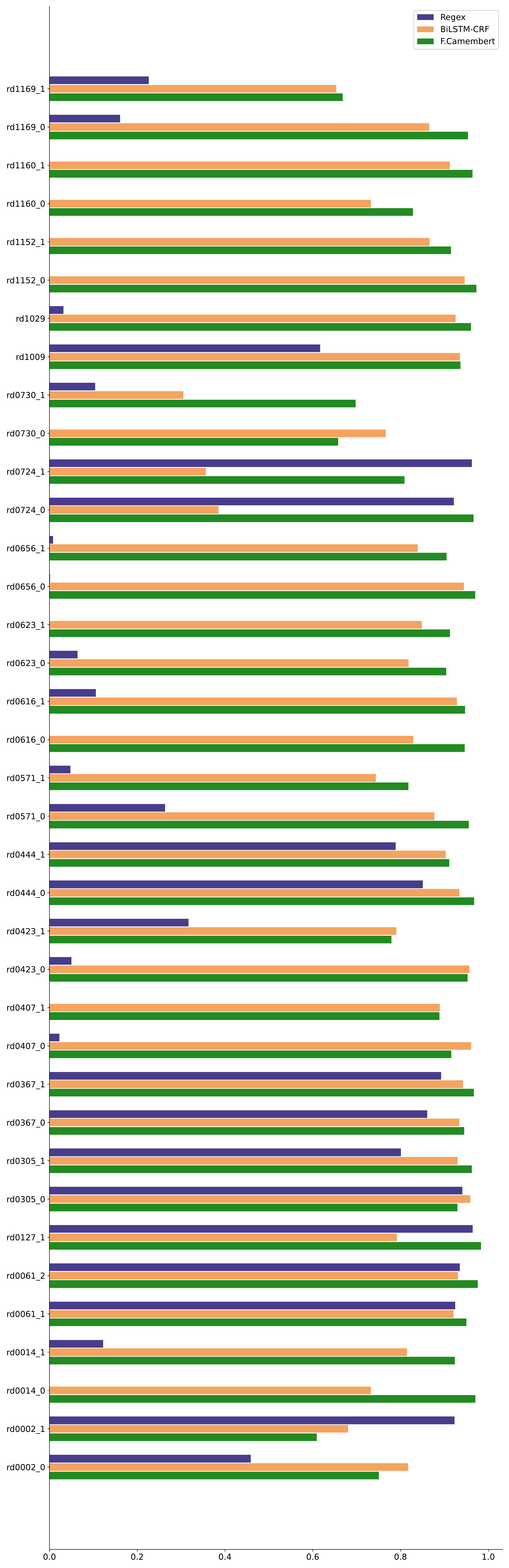}
         \caption{Token-level F1}
     \end{subfigure}
     \hfill
     \begin{subfigure}[b]{0.4\textwidth}
         \centering
         \includegraphics[width=\textwidth]{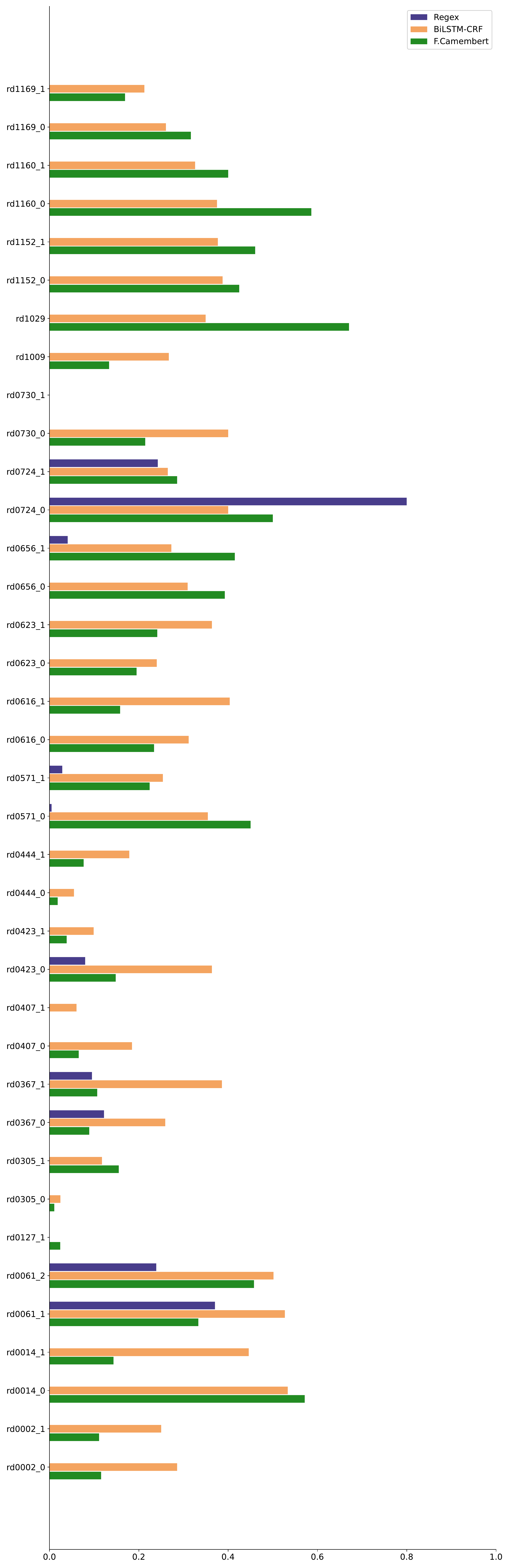}
         \caption{SSM F1}
     \end{subfigure}
        \caption{Results per file from Test$_{N}$.}
        \label{fig:ood_per_file}
\end{small}
\end{figure*}

Figure \ref{fig:ood_per_file} shows the results obtained by the considered baselines on each file from Test$_{N}$, the out-of-distribution corpus.

\section{Computing information}
\label{app:compinfo}
We trained the models on a 32-core Intel Xeon Gold 1051 6134 CPU @ 3.20GHz CPU with 128GB RAM 1052
equipped with 4 GTX 1080 GPUs with 11GB RAM 1053
each.
The required time for train, where applicable, validation and test both on Test$_{C}$ and Test$_{N}$ was a bit less than 2 hours: 15 minutes for \textit{Regex}, 45 minutes for \textit{\textit{BiLSTM-CRF}}, and 40 minutes for \textit{CamemBERT}.

\section{Clause-consistent predictions} \label{sec:clause-consistent}

We lead a final post-processing experiment on top of the predictions made by the different models.

This step is meant to ensure the consistency of the automatic annotations at the {clause}\footnote{Here, the term \textit{clause} does not strictly follow its grammatical definition, but it is used to designate any sequence of words between two punctuation marks.} level. 
It relies on a simple heuristic drawn from the knowledge of the task: all words between two consecutive punctuation marks (full stop, question mark, hyphen, quotation mark, etc.) lie at the same narrative level, ie. the sequence of words is either uttered by a character or part of the narrator's discourse. Thus, all words stemming from a common clause must be associated with the same label.

In practice, this is implemented as a post-processing step.
Based on a model's predictions, the clause-consistency is ensured by imposing all words from the same clause to be associated with the same label. For each clause, a majority vote is carried out from the predicted labels to determine a consistent unique label for all the words of the clause.

\begin{table*}
\begin{small}
\centering
\begin{tabular}{r|ccc|ccc}
\hline
& & \textbf{Test$_{C}$} & & & \textbf{Test$_{N}$} & \\
& \textit{Regex} & \textit{BiLSTM-CRF} & \textit{F.CamemBERT} & \textit{Regex} & \textit{BiLSTM-CRF} & \textit{F.CamemBERT} \\
\hline
Tok. F1 & 90 & 83 & \textbf{96} & 47 & 88 & \textbf{93}\\
 & = & = & = & = & = & =\\  & & & & & & \\
SSM F1 & 45 & 73 & \textbf{81} & 5.5 & \textbf{34} & \textbf{34}\\
 & = & +1 & +5 & = & +4 & +8\\ & & & & & & \\
ZME & 0.23 & 0.41 & \textbf{0.09} & 1.09 & 0.29 & \textbf{0.21}\\
 & = & = & = & = & = & -0.03\\ & & & & & & \\
\hline
Av. Tok. F1 & 90 (2.3) & 84 (20) & \textbf{95} (3.8) & 36 (39) & 82 (16) & \textbf{90} (10) \\
 & = & = & = & = & = & +1 \\ & & & & & & \\
Av. SSM F1 & 43 (17) & 72 (21) & \textbf{78} (13) & 5.5 (15) & \textbf{33} (15) & 31 (22) \\
 & = & +1 & +6 & = & +5 & +7 \\ & & & & & & \\
Av. ZME & 0.24 (0.05) & 0.52 (0.85) & \textbf{0.11} (0.08) & 1.13 (1.06) & 0.54 (0.79) & \textbf{0.28} (0.23) \\
 & = & = & = & = & -0.01 & -0.02 \\
\hline
\end{tabular}
\caption{Results after clause-consistent (CC) post processing, overall (top) and averaged over files (bottom) with standard deviations in parentheses on clean ($C$), and noisy ($N$) test-sets. Best scores are in bold. These scores can be put into perspectives with regards to performances of the models without CC post-processing, as displayed in~\autoref{tab:whole_results}. For each score, the second row shows the absolute difference between the scores obtained from the post-processed and raw predictions.} \label{tab:whole_results_post_cc}
\end{small}
\end{table*}

Results of the clause-consistent (CC) post-processing experiment are disclosed in~\autoref{tab:whole_results_post_cc}.
As expected, this post-processing step has only few to no impact on the \textit{Regex} model's output. Indeed, this method directly labels sequences of words caught with regular expressions that are redundant with the definition of clauses. However, imposing clause-consistent predictions allows to significantly improve the performances of the \textit{BiLSTM-CRF} and \textit{F.CamemBERT} based models. 
Overall, this heuristic never deteriorates the performances of the models on all performance scores, and the improvements are particularly significant for sequence-level metrics.

Enhancements are striking for the \textit{F.CamemBERT} in all evaluation configurations. This post-processing step allows to alleviate one of the observed weaknesses of the model (see \autoref{subsec:qualitative}) by hindering sequences of alternated labels within the same clause. This results in major performance boosts of up to 8 points for the overall SSM F1 on Test$_N$. On the other hand, sequence level performances of \textit{BiLSTM-CRF} also benefit from the CC-predictions but mainly on noisy files with a gain of 4 points on the overall SSM F1.

Clause consistent predictions allow to reach fairly high scores even on the most challenging task of strict sequence match on well-formated documents, \textit{F.CamemBERT} reaching on average a SSM F1 score of 78 on Test$_C$. Yet, performances on noisy files remain curtailed \textit{F.CamemBERT} and \textit{BiLSTM-CRF} SSM F1 scores on Test$_N$ are on average, respectively, 31 and 33 with large variances among files.

\end{document}